\documentclass[twocolumn]{article}
\usepackage{fullpage}
\usepackage{moreverb,url}
\usepackage{siunitx}
\usepackage{authblk}
\usepackage[colorlinks,bookmarksopen,bookmarksnumbered,citecolor=red,urlcolor=red]{hyperref}
\usepackage{amsmath, amssymb, bm}
\usepackage[utf8]{inputenc}
\usepackage{graphicx}
\graphicspath{{}{fig/}}
\usepackage{float}
\usepackage{subcaption}
\usepackage{xcolor}
\usepackage{placeins}

\usepackage[]{natbib}
\newcommand{\concept}[1]{``{\textit{#1}}''}
\newcommand{\meth}[1]{\textbf{#1}}
\newcommand{\SUDS}{\meth{SUDS}}
\newcommand{\TLS}{\meth{TLS}}
\newcommand{\GMR}{\meth{GMR}}
\newcommand{\GMRS}{\GMR+\SUDS}
\newcommand{\TLSS}{\TLS+\SUDS}

\begin{document}

\title{Sample Efficient Learning of Body-Environment Interaction of an Under-Actuated System}
\author[1]{Zvi Chapnik}
\author[1]{Yizhar Or}
\author[2]{Shai Revzen\thanks{Corresponding author: Shai Revzen, EECS Room \#4225, 1301 Beal Ave, Ann Arbor, MI 48109, USA \texttt{shrevzen@umich.edu}}}

\affil[1]{Mechanical Engineering, Technion, Haifa, Israel}
\affil[2]{Electrical Engineering and Computer Science, Ecology and Evolutionary Biology, University of Michigan, Ann Arbor, MI USA}

% \keywords{Geometric Mechanics, Learning, Under-actuated system}
\maketitle

\tableofcontents
\begin{abstract}
Geometric mechanics provides valuable insights into how biological and robotic systems use changes in shape to move by mechanically interacting with their environment.
In high-friction environments it provides that the entire interaction is captured by the ``motility map''.
Here we compare methods for learning the motility map from motion tracking data of a physical robot created specifically to test these methods by having under-actuated degrees of freedom and a hard to model interaction with its substrate.
We compared four modeling approaches in terms of their ability to predict body velocity from shape change within the same gait, across gaits, and across speeds. 
Our results show a trade-off between simpler methods which are superior on small training datasets, and more sophisticated methods, which are superior when more training data is available. 
\end{abstract}
\section{Introduction}
With the exception of reaction-mass systems like rockets, virtually all locomotion propulsion is accomplished through shape-change that produces reaction forces from the environment to propel the body.
In most animals, when they move quickly, and in virtually all man-made devices, these shape changes are \concept{rhythmic}\footnote{As distinct from \concept{periodic}, which is a stronger requirement.} --- they follow nearly the same sequence of body shapes, at a frequency which changes little cycle to cycle.
When the repeating shape change is pronounced, repeating sequence is often called a \concept{(periodic) gait}.
A vast literature exists attempting to relate body shape changes to body motion: turning of wheels to motion of cars; flapping of wings to flight of birds or flies; spinning of blades to propulsion of submarines or of helicopters; cycling of legs to running or walking, etc.
Many moving biological and robotic systems have passively deforming mechanical parts, most obvious among these in biology are flapping fins, bending wings, and beating flagella.

In systems where momentum decays quickly, such as small bodies swimming in high-viscosity fluids \citep{purcell1977talk, shapere1989geometry}, or walking with many legs \citep{zhao2021locomotion}, or more generally in cases where momentum decays quickly enough for any other reason \citep{kvalheim2019gait}, these theories of propulsion take on a particularly simple form called the \concept{motility map} and arising from \concept{geometric mechanics}.

The contributions of our work here are:
\textbf{(1)} a machine learning model especially suited to the motility map of an under-actuated system, and \textbf{(2)} a study of the sample efficacy of this method and its comparison with other methods, applied to multiple gaits of a physical robot.

\subsection{Geometric Mechanics \& Motility Maps}

In 1977 Purcell gave a notable talk, discussing ``life in low Reynold numbers'' \citep{purcell1977talk} and expressing how the physics of propulsion is fundamentally different for small organisms.
The interest in this question in the physics community lead Shapere and Wilczek to publish a paper in 1989 that arguably started the field of geometric mechanics by giving a general theory of how shape-change in microorganisms produced motion \citep{shapere1989geometry}.
They showed that in the low Reynolds number fluid dynamics regime that governs microorganisms there appears a first-order equation of motion which is linear in the rate of shape change.
This is quite unlike Newtonian dynamics that are typically used for modeling, and which take the form of second order ODE-s.
It is also \concept{geometric} or \concept{principally kinematic} in the sense that only the shape of the motion matters, but the rate of motion does not -- changing the body shape twice as fast moves the body through the world exactly twice as fast along the same trajectory.
The translation and rotation created by a motion depends only on the sequence of shapes. 

The modern treatment of Shapere and Wilczek's result derives its mathematical form from work done in the 90s by Bloch, Marsden, Ostrowski, and others under the name of ``geometric mechanics''.
We direct the interested reader to the brief introduction to geometric mechanics in \citet{hu2025learning},  the treatis \citet{marsden1994introduction}, or the up-coming book \href{https://www.rosslhatton.com/GeometryOfMotion.html}{The Geometry of Motion} by Hatton.

Geometric mechancis shows how to decompose symmetric (Lagrangian) dynamics into \concept{body frame} and \concept{shape} coordinates. 
The key result of geometric mechanics that is of interest here is the Reconstruction Equation \cite[eqn.1.2.1]{bloch1996nonholonomic}, describing the body velocity $v_b$ in terms of the body shape $r$, its rate of change $\dot r$, and a ``group'' momentum $p$ associated with the moving body frame
\begin{align}
  v_b = A(r) \dot r + \mathbb{I}^{-1}(r) p. \label{eqn:recon}
\end{align}
The group momentum affects body velocity via the shape dependent \concept{frozen inertia tensor} $\mathbb{I}(r)$, and the shape velocity affects the body velocity via the \concept{(local) connection}, also called the \concept{motility map}.

For many systems, $p=0$, or $p \to 0$ so quickly that the $p = 0$ assumption is reasonable, leaving only $A(r)$ to govern the motion; such systems are called ``principally kinematic'', or ``geometric''.

\subsection{Optimal Gaits for 2-dimensional Shape-Spaces}

While the mathematical elegance of the geometric mechanics theory of the 1990-s was appealing, the real breakthrough in understanding gaits came through the work of Hatton and Choset, more notably \citet{hatton2011geometricMP}. 
That paper resolved a key technical issue arising from the fact that the \concept{body velocity integral (BVI)}, which is the component-wise integral of the body velocity $v_b$, does not accurately reflect how the body has moved, because, e.g. early rotations affect the direction of later translations, but the BVI does not capture this.
Here \citet{hatton2011geometricMP} showed how to choose a body frame as a function of shape such that the difference between BVI and the actual motion given by the \concept{path ordered integral} of $v_b$ is minimized.
They then showed how, for 2-dimensional body shapes, Stokes' theorem could be used to directly find maximal translation gaits as 0-level sets of a \concept{height function} that arises from the motility map.

\subsection{2D Data Driven Motility Maps}

With the realization that motility maps could be used to derive optimal gait came the hope that motility maps could be obtained by directly measuring robot-environment interactions.
This was first done in \citet{hatton2013geometric}, which obtained the motility map of a 3-segment Purcell-swimmer-like robot moving in a granular fluid, and used this to design maximal displacement gaits for this swimmer.
Unfortunately, even for a 2-DoF, 3-segment swimmer finding the motility map required sampling the 4-dimension $r \times \dot r$ space, making this method extremely challenging in practice.
Even today, most groups' work on optimal gaits relies on reducing the shape-space to 2 ``modes'' and using Hatton's height-function methods \citep{chong2021Coordination, chong2022General, chong2023GaitRSS}.

\subsection{High Dimensional Data Driven Motility Maps}
The 2D limitation for physical systems was circumvented in 2018 when Bittner, Hatton, and Revzen proposed an alternative \citep{bittner2018geometrically}.
By applying methods of \concept{Data Driven Floquet Analysis (DDFA)} \citep{revzen2015data, revzen2017locomotion}, they obtained a model of the motility map restricted to a tube surrounding a gait.
They modeled the gait itself as a periodic function of phase using a Fourier series, and expanded the motility map in a first order Taylor series around the nominal gait.
In their method, each data sample was first assigned a phase $\phi$.
They binned the $\phi$, producing a bin index $k$, and built the following model, estimating $(v_b)^k_n$, the $n$-th sample of body velocity at time index $n$, from $\delta_n$ and $\dot \delta_n$, the shape and shape velocity offsets from the gait cycle,  
\begin{align}
    (v_b)_n^k \sim& \left(\sum_i B^k_i\delta ^i_n\right)+\left(\sum_i A^{k}_i \dot{\delta}^{i}_n\right)\nonumber\\
    &+ \left(\sum_{i,j} \frac{\partial A^{k}_i}{\partial r_j} \delta^{j}_n \dot{\delta}^{i}_n\right)+ C^k.
\end{align}
The then interpolated the tensors $C^k$, $B_i^k$, $A_i^k$ and $\frac{\partial}{\partial r_j}A^k_i$ using Fourier series.
Since this a linear model for each phase, they could reliably build a model using a number of cycles that is at most quadratic in $\dim r$ --- a breakthrough improvement when compared with the \citet{hatton2013geometric} approach which was exponential in $\dim r$.
While a single such sampling did not give an optimal gait, it could be used to make a gradient descent step in an optimization, and Bittner showed how optimal gaits with hundreds of parameters could be learned on both simulated and physical robots with about as many cycles of data.
The \concept{(phase based) Total Least Squares (\TLS)} method used in \citet{bittner2018geometrically} is one of the methods we compare here.

\subsection{Modeling Underactuated Systems}
In a follow-up paper \cite{bittner2022data}, Bittner and his collaborators considered that many biological and robotic systems have passively deforming mechanical parts.
The most obvious among these in biology are flapping fins, bending wings, and beating flagella.

The \citet{bittner2022data} paper divided the shape variables into $r=(r_a,r_p)$ , where $r_a$ are actuated shape variables and $r_p$ are passive shape variables, and showed that if the passive shape variables $r_p$ are governed by a fairly general visco-elastic interaction term with the actuated shape variables $r_a$ and the with the environment (see also \citet{murray2017mathematical}), the dynamics can be put in a form very similar to the motility map:
\begin{align}
    v_b &= A(r) \dot r_a +B(r) \nonumber\\ 
    \dot r_p &=C(r)\dot r_a+D(r).
    \label{eq:conUA}
\end{align}
Systems with this structure are called \concept{Shape Underactuated Dissipative Systems (SUDS)}.
In Eqn.~\ref{eq:conUA}, $A(r)=A(r_a,r_p)$ is the familiar motility map; $B(r)$ a shape-dependent bias term; and $C(r)$ and $D(r)$ an affine map from the actuated variables to the passive ones.
Notably, the dimension of $C$ is $\dim(r_p)\times\dim(r_a)$, which is potentially much smaller than $\dim(r)\times\dim(r)$.
The model itself is still linear and therefore just as easy to learn with  phase dependent total-least-squares.
We use this model as one of the models in our comparison, referring to it as \TLSS.
\citet{bittner2022data} showed the efficacy of \TLSS\ on simulated systems, but never tested the method on a physical robot.
One of the contributions of the current work is its place as the first such experimental test.

\subsection{Augmented Gaussian Branching Regression}

In 2025, \citep{hu2025learning} proposed a specific sample-efficient machine learning method specially tuned to learning motility maps, called \concept{Augmented Gaussian Branching Regression (A-GBR)}.
Their method was able to learn the motility maps of fully actuated robots with about $\times 10$ fewer data samples than a multi-layer-perceptron. 
Furthermore, their method did not require the input data to be from rhythmic motions with a well-defined phase.
They expanded upon \concept{Gaussian Mixture Regression (GMR)} \citep{sung2004gaussian}, a type of Nadaraya-Watson estimator \citep{nadaraya1964estimating, watson1964smooth}, to build a model of a generalized motility map
\begin{align}
    v_b = A(r_a,\dot r_a), \label{eq:genmot}
\end{align}
adding the capability to be multi-valued based on recent history (the ``branching''), and having hyper-parameters which bias the method towards linearity in $\dot r_a$ (the ``augmentation'').
We use $r_a$ rather than $r$ to remind the reader that this model uses only part of the state -- the actuated shape variables.

Because their method can be just as easily applied to the fully actuated case $v_b = A(r_a, \dot r_a)$ and to the under-actuated case $(v_b, \dot r_p) = A(r_a,r_p,\dot r_a)$, (the $\dot r_p$ being provably unnecessary in the RHS; see \citet{bittner2022data}) we used it in both forms.
These lead to the methods called \GMR\ and \GMRS\ below.
More generally, we use the term ``adding \SUDS'' to indicate the process of modifying a model to include the passive shape variables $r_p$ and their dynamics. 

% \SR{Another note: one of our contributions here is the use of PCA to identify the significant passive DoF - this should be in intro where we state our contribution, and discussed in the discussion}

\section{Methods}

To test the ability of various methods to identify the motility map of a SUDS, we built a system which would have the following properties
\begin{description}
\item[Friction dominated] The system's motion was so friction dominated that it effectively had no momentum -- it stopped in place the moment its shape stopped changing.
\item[Under-actuated] The system had 4 un-actuated shape variables, and only 2 actuated shape variables, i.e. it was ``mostly'' un-actuated.  
Yet these un-actuated shape variables could meaningfully affect its motion.
\item[``Impossible'' to model] We chose a system for which no existing physics models could reasonably be formed, to highlight the power of a data-driven / learning approach.
\item[Directly observable shape] All the shape variables of the system could be measured in a straight-forward manner.
\end{description}

\subsection{A 4-Flipper, 3-Segment Granular Swimmer}

The system we selected is similar to the 3-segment swimmers familiar from many classical papers in geometric mechanics \citep{purcell1977talk, hatton2013geometric, zigelman2024purcell, yona2019wheeled, hatton2013swimming, shammas2007geometric, rizyaev2024locomotion, melli2006motion} -- it is a 3-segment swimmer moving through a granular material. 
We chose red beans as the grains, for several reasons: (1) Beans are non-spherical, and it is only in the last couple of years that a kinetic theory of granular materials with non-spherical grains is beginning to form \citep{virozub2019planar, hanif2024dynamical}; (2) Red beans were cheaply available in bulk; (3) Because beans are large enough, it was easy to seal the robot against beans entering the mechanism and causing jamming.

\begin{figure*}[h]
    \centering
    \includegraphics[width=0.45\linewidth]{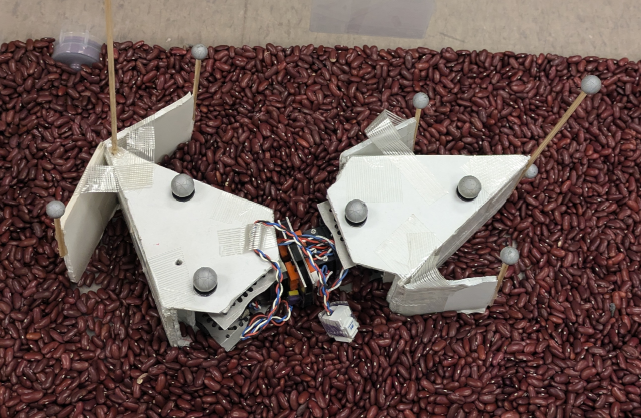}
    \hspace{3mm}
    \includegraphics[width=0.45\linewidth]{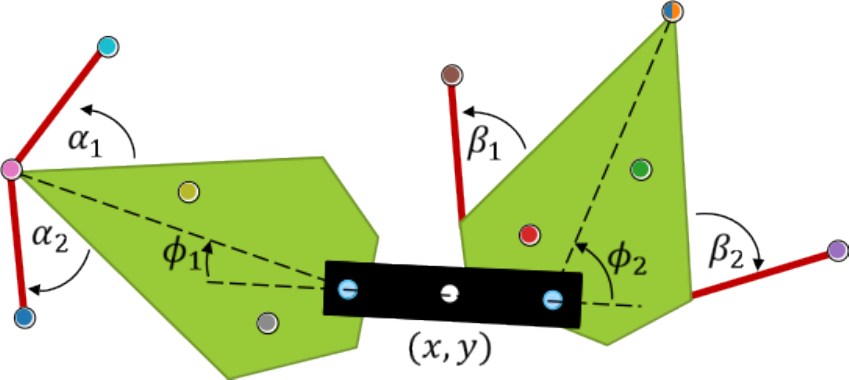}
    \caption{4-flipper, 3-segment swimmer. %
      Picture of the swimmer in the granular medium (left), and schematic showing markers and angles (right). %
      If the robot axes were all perfect hinge joints, the shape of the robot was fully characterized by the two motor angles $\phi_1,\phi_2$, and the flippers angles $\alpha_{1,2},\beta_{1,2}$. %
      We measured the swimmer’s pose using 11 markers: 4 on the tail link --- two on its top face (red, green dots), one raised on a spoke, and one lowered and attached to the back edge (teal, orange half-circles); 3 on the head link -- two on the top face (yellow, gray), and one raised (magenta); one each on each of the flippers (gunmetal blue, cyan, purple, and brown). }
    \label{fig:pic and sketch}
\end{figure*}

To this classical, 2-actuator, 3-segment swimmer design, we added 4 passive ``flippers'' attached to the body on live hinges made out of fiber reinforced tape (Scotch \#8959). 
Each flipper was limited by a piece of fiber tape allowing it to open only up to about $80^\circ$ away from the body surface to which it was attached (see Fig.~\ref{fig:pic and sketch}).

\subsection{Measurement Instrumentation}

To make it easy to observer the swimmer state, we made the swimmer frame substantially higher than the depth to which it sinks in the granular bed. We also added motion tracking markers (obtained from the motion tracking system vendor) and attached them to rigid spokes (bamboo skewers) which we hot glued and taped to the robot frame (see Fig.~\ref{fig:pic and sketch}).

For all of the experiments we used 4 Qualisys Oqus-310+ cameras, running at 100 [Hz] with QTM 2.17 build 4000 software, recorded with the QTM gap-filling disabled.

This marker-based measurement system gave us the spatial location of each marker, to sub-millimeter precision according to the vendor's calibration results.
From these data we derived both body frame and shape of the swimmer.

\subsection{Data Processing}

We scanned the marker tracks for jumps larger than 9 mm.
Less then 0.1 \% of the data points were adjacent to such a jump. 
We removed those points, and filled the gaps by linear interpolation.

We chose a body frame as follows.
We chose its XY plane to be the least-squares fit to the plane containing the 4 top-surface markers, and maintaining a constant Z offset from the remaining segment markers, i.e. it is a linear combination of segment markers that minimizes the Z variance, and has the mean Z of the top-surface markers equal to 0.
We chose origins for two segment frames to satisfy (in a least-squares sense) the conditions that: (1) they are in the XY plane; (2) the origins maintain a constant distance of each other.
Condition (1) is an arbitrary choice; condition (2) expresses the fact that the segments were rigidly connected to each other, and we chose points on the motor axes' intersection with the XY plane as the segment origins.

We took the mid-point of these origins as the origin of the body frame, with the line connecting the segment origins as the X axis, thereby fully defining the body frame.
%We did that by finding the least square of the high value of each marker minus the known values.
%By that we found the orthogonal axis in each time frame as a linear combination of the markers values.
%The next step was to find the axis point of each links' motor, we used least square to find those two points as a linear combination of the base markers, to find constant length between them.
%Now, we were able to calculate the actuated angles $\phi_1,\phi_2$, also we were able to define a rotating coordinates framework that rotated with the swimmer.
%In order to use the four modeling methods we defined the actuated DoF as the mean value of the base markers movement in each link about the axis motor point of the link- defining $r_a$ in position [mm] and not in angle [rad].
%For the under actuated DoF we had measurements of the four flippers location and the two high leveled base markers, we looked only on the x-y part of the measurement in the rotating frame work.

We then projected the two top-surface markers of each segment on the XY plane, and took the their (2D) centroid as representing the pose of the segments relative to the body, and defining the segment's X-axis.  
We transformed all 4 flipper markers into the segment coordinate frames of the segments they were connected to, taking those positions to represent the shape of the flippers.
This allowed for the possibility that flipper axes were not perfectly aligned with the Z axis of the body.
We used cartesian values rather than angles to represent the pose, to avoid the heteroscedasticity caused by applying inverse trigonometric functions to cartesian marker locations.

% !!! proj XY with upper markers and oscillation 6x2
When inspecting our data, we noted that an unexpected parasitic oscillation appeared -- the markers raised on spokes had enough inertia to oscillate.
We therefore combined the XY projections of the two markers raised on spokes and the XY projections of the flipper markers into a 12 dimensional state representing the un-actuated state. 
We omitted the Z coordinates of these markers because they seemed to move so little in Z as to have an insignificant systematic structure.
We then reduced the 12-dimensional passive shape down in dimension by applying Principal Component Analysis (PCA).
We took the 4 largest components as the ``passive'' shape variables $r_p$.
These 4 PC-s captured 96\% of the variance in the original 12 dimensional un-actuated state data.

\subsection{Actuation and Gaits}

\begin{figure*}[h]
    \centering
    \includegraphics[width=0.32\linewidth]{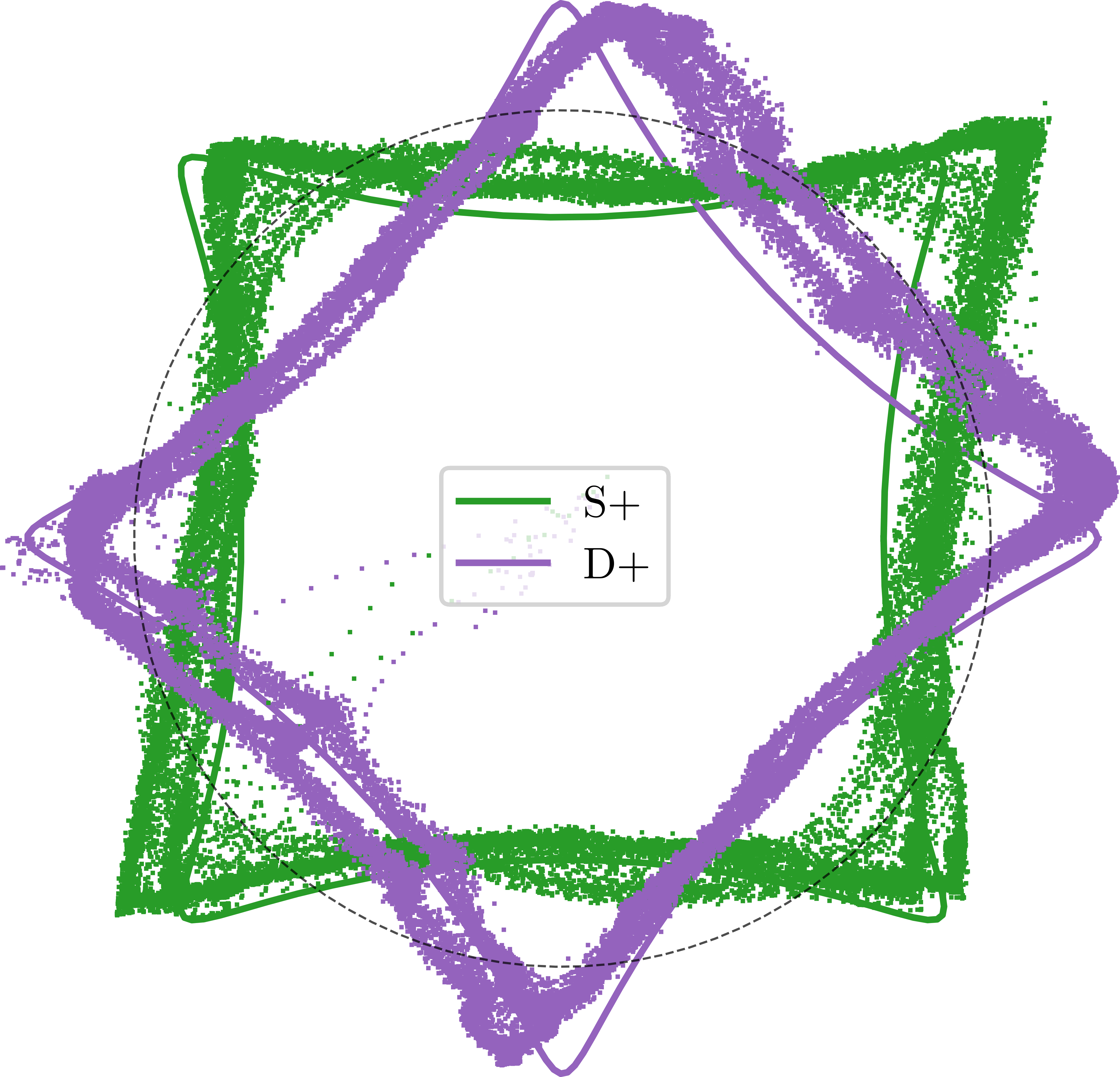}
    \hspace{3mm}
    \includegraphics[width=0.32\linewidth]{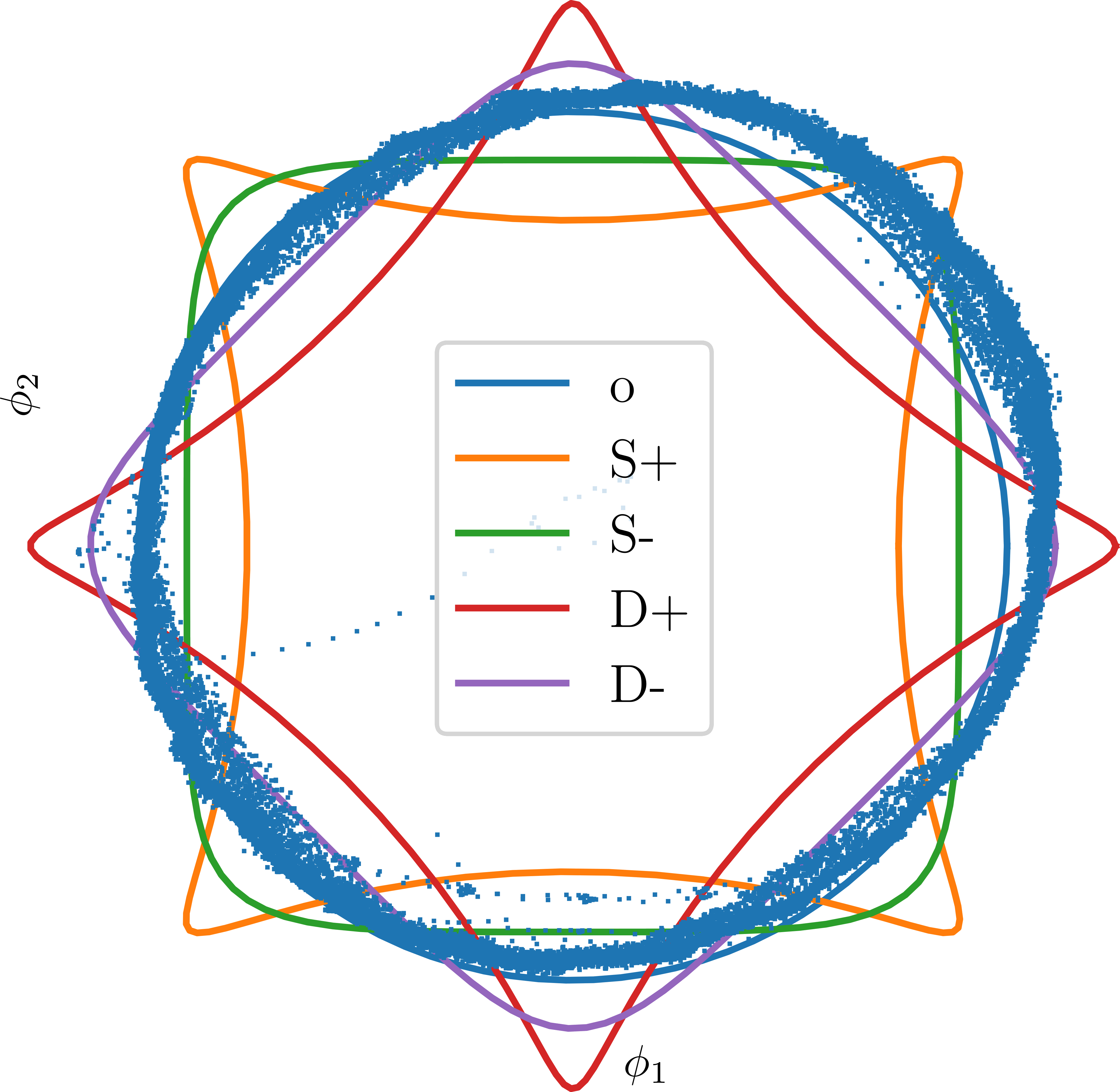}
    \includegraphics[width=0.32\linewidth]{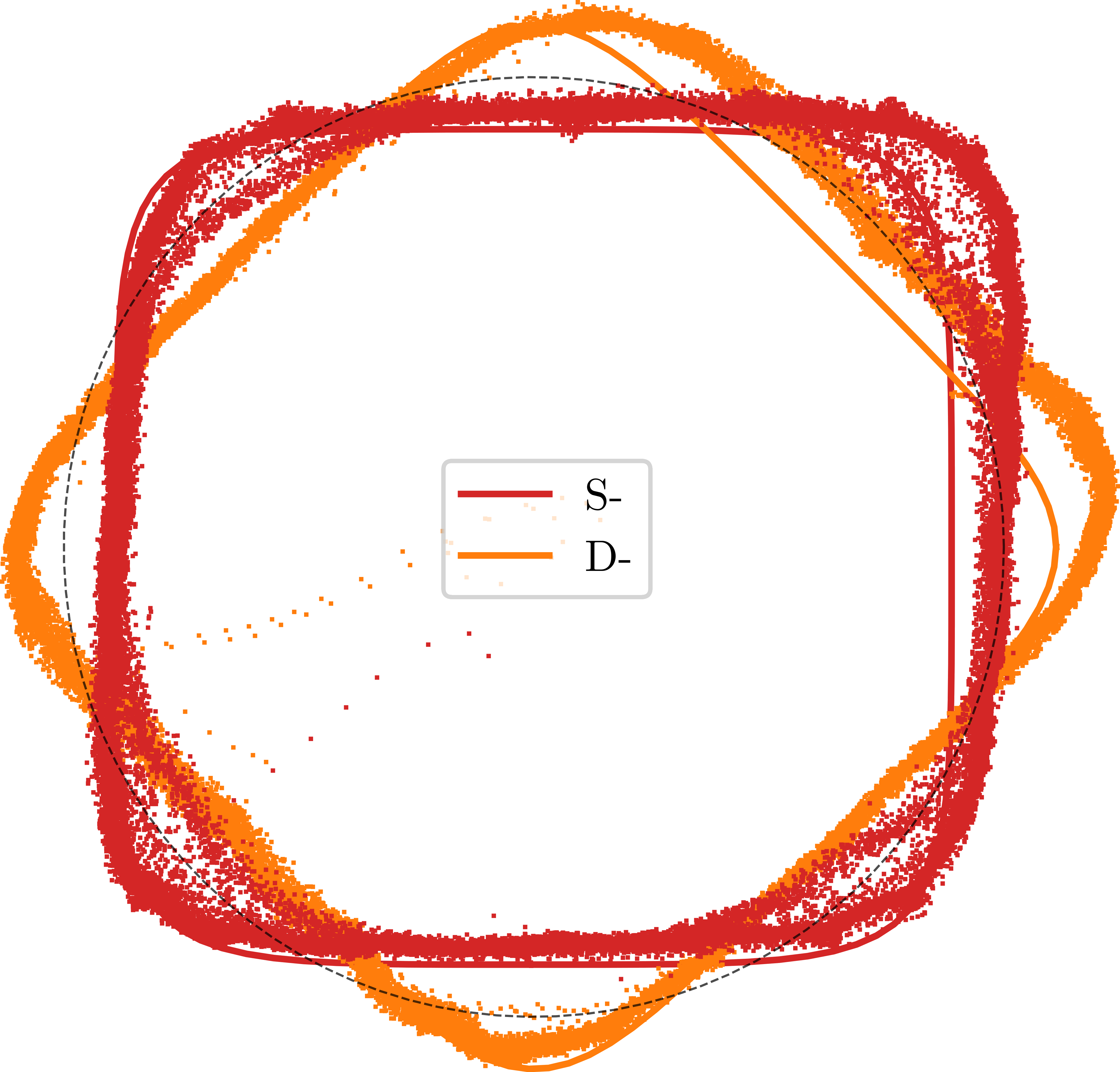}
    \caption{Commanded and measured theoretical gaits. %
    We plotted the commanded $(\phi_1, \phi_2)$ values (solid lines) and the measured angles (dots, same color). %
    For scale, we included a circle of radius 1 radian (dashed black). Mild diamond and square (D-,S-) built by: $\phi_1=\cos(\theta)\pm\cos(3\theta)/9,\phi_2=\sin(\theta)\mp\sin(3\theta)/9$, and extreme diamond and square (D+,S+) built by: $\phi_1=\cos(\theta)\pm\cos(3\theta)/4,\phi_2=\sin(\theta)\mp\sin(3\theta)/4$ }
    \label{fig:gaits}
\end{figure*}

We actuated the swimmer in 5 different periodic gaits, representing a circle (O), mild and extreme diamond shape (D-, D+), and mild and extreme square shape (S-, S+), all of average amplitude 1 radian (see Fig.~\ref{fig:gaits}), given as Fourier series.
For each gait, we ran the swimmer at 3 frequencies: $1/3$,$1/2$, and $1$ Hz.
Between trials we remixed and leveled the beans in the tub, and reset the swimmer to approximately the same location.
We then ran the swimmer for 10-30 cycles -- until it came close to running out of room.
We repeated each condition at least 4 times.

\subsection{Learning}

We compared four of models for learning the motility map, arranged along two axes: \TLS\ vs. \GMR, and with or without a model of the passive shape variables: \TLS, \GMR vs. \TLSS, \GMRS.
The \TLS\ models used the phase based total least squares methods of \citet{bittner2018geometrically, bittner2022data}, whereas the \GMR\ methods applied the A-GBR algorithm of \citet{hu2025learning}.

We evaluated each model using a bootstrap process. 
In each bootstrap iteration we trained and tested each model with the same number of data points.
We used 30 cycles to train each model by picking a trial at random, and a 30 cycle interval from that trial. 
We down-sampled all frequencies to have the same number of data points by direct decimation (no down-sampling filter applied), so as to preserve identical measurement error characteristics.
We then tested the models against a different trial that was randomly selected using the same process.
We repeated this bootstrap process for each model, for all pairs of gait and frequency against all pairs of gait and frequency.

The ``RMS error'' we report as our goodness of prediction metric is the RMS prediction error of the robot's velocity along its own X axis, in arbitrary (but consistent) units.
While other components of the motion -- Y velocity, rotation rate, and even changes in Z -- existed, they were very small and noisy, and thus less suitable for generating a comparison between models.

\section{Results}

The data-frame of our results was an array of $4 \times (5 \times 3) \times (5 \times 3) \times 47,000$ --- 4 models, 5 gaits, 4 frequencies, and $47,000$ samples at each condition.
To interpret these results we classified pairings of train and test on the frequency similarity and gait similarity axes, allowing us to examine both in-sample learning and out-of-sample generalization.

\subsection{Log-RMS Error Distributions}
\begin{figure}[h]
    \centering
    \includegraphics[width=1\linewidth]{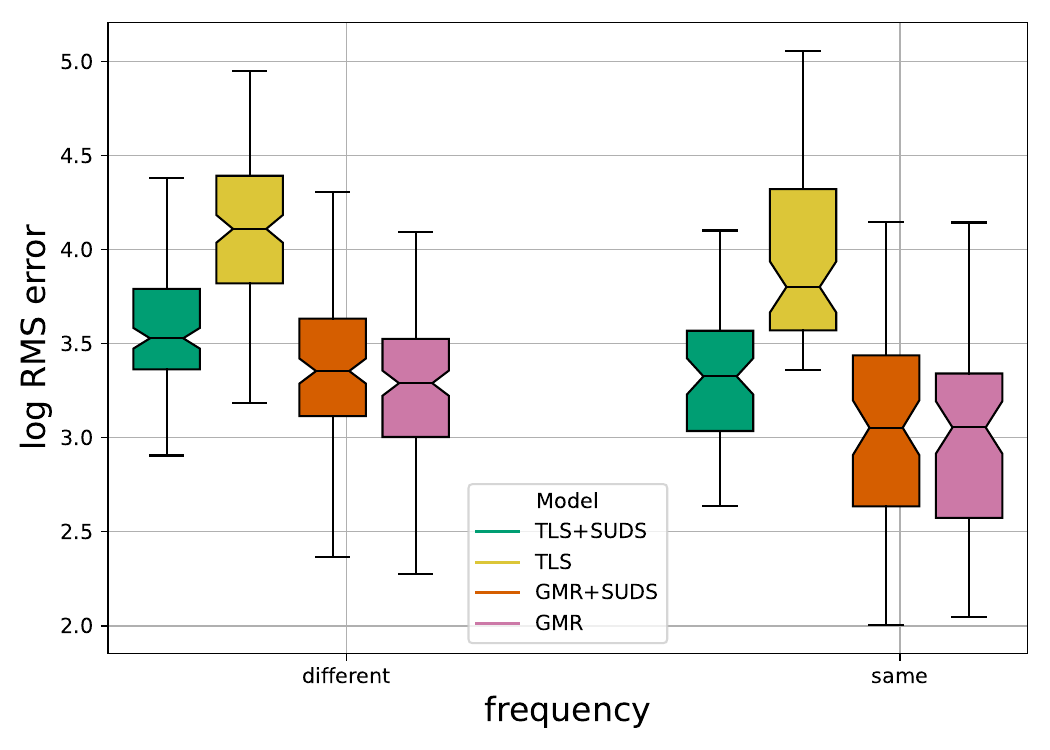}
    \caption{Prediction error by frequency similarity and model type. %
    We plotted box plots (default \texttt{matplotlib} parameters) of each category: same frequency error distributions (right), and different frequency error distributions (left).}
    \label{fig:freq}
\end{figure}

On the frequency axis, we categorized parings as same-frequency and different-frequency (see Fig.~\ref{fig:freq}).

\begin{figure*}[h]
    \centering
        \includegraphics[width=1\linewidth]{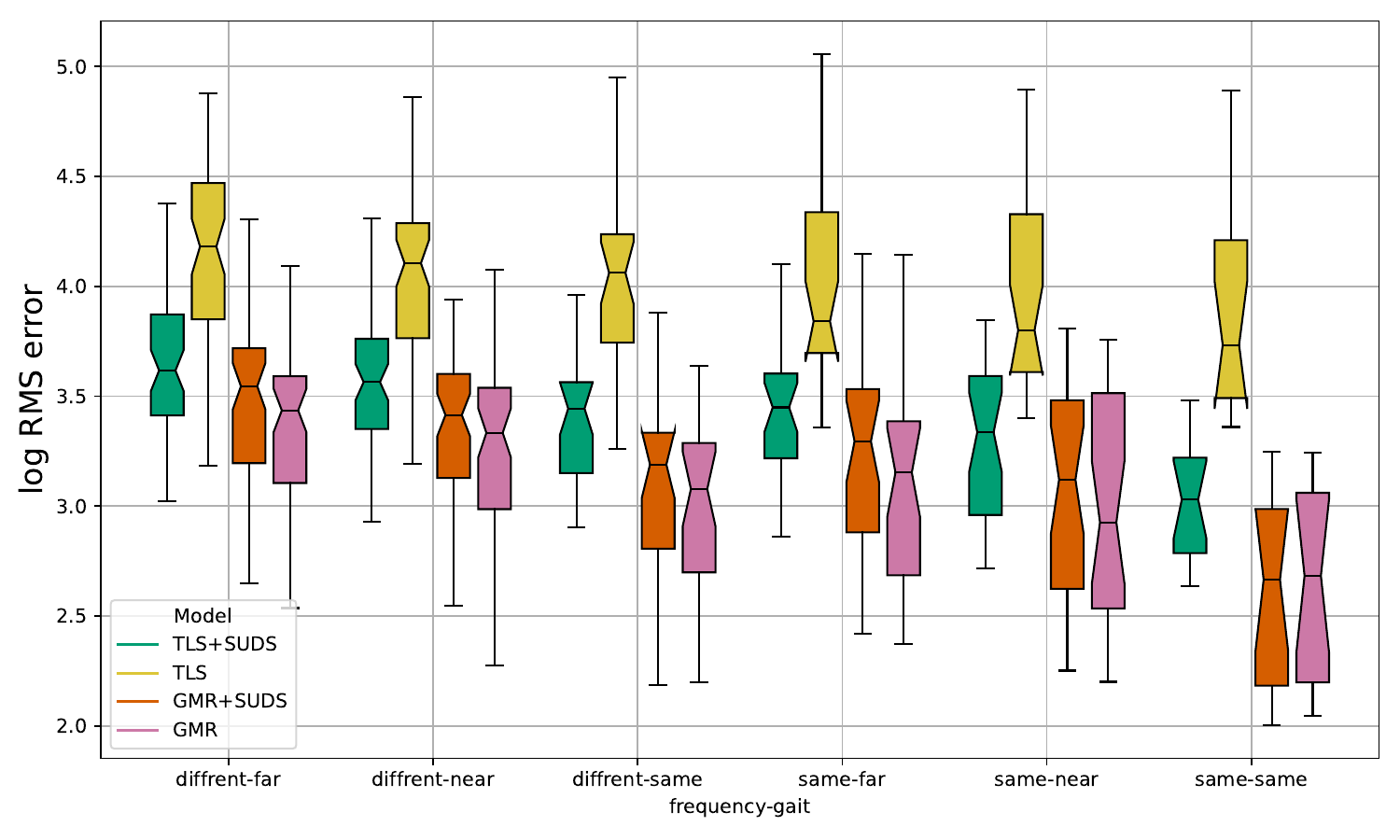}
    \caption{Prediction error by combined frequency similarity and gait similarity, by model type. %
    We plotted box plots (default \texttt{matplotlib} parameters) of log RMS error distributions each category.}
    \label{fig:both}
\end{figure*}

On the gait axis running S+,S-,O,D-,D+, we categorized pairing as same-gait, near-gait -- one step apart, and far-gait -- more than one step apart (see Fig.~\ref{fig:gaits}).
\begin{figure}[h]
    \centering
    \includegraphics[width=1\linewidth]{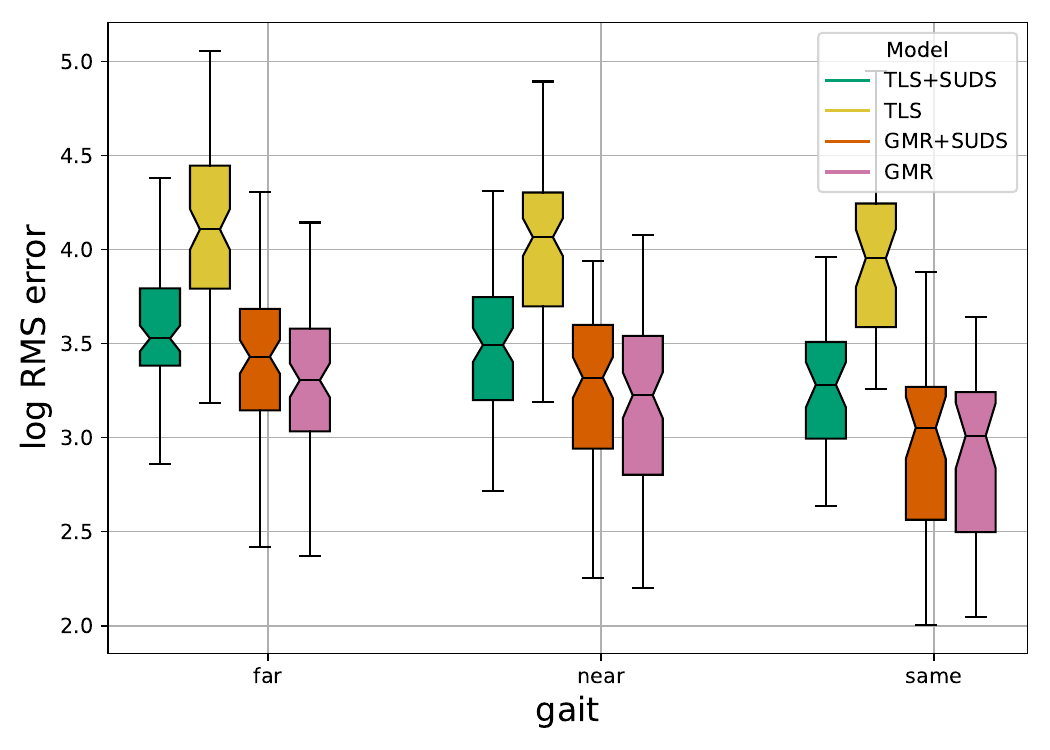}
    \caption{Prediction error by gait similarity and model type. %
    We plotted box plots (default \texttt{matplotlib} parameters) of log RMS error distributions each category: same gait (right), near gait (middle), and far gait (left).}
    \label{fig:geo}
\end{figure}
We also considered the possibility of frequency and gait geometry interactions (see Fig.~\ref{fig:both}).

In Figure \ref{fig:freq} we showed the log of the RMS prediction error for each model for same and different frequencies for all of the geometries.
For both, same and different frequencies the \GMR\ based models increased the prediction accuracy approximately 40\% from the \TLS\ based models.
Adding the \SUDS\ to the \GMR\ in this case increased the accuracy by around 3\%.
The \TLSS\ based models increased the prediction accuracy approximately 30\% from the \TLS\ based models.

In Figure \ref{fig:geo} we showed the log of the RMS prediction error for each model for same and near and far geometries for all of the frequencies.
For same geometry the \GMR\ based models increased the prediction accuracy by 48\% from the \TLS\ based models.
The \TLSS\ based models increased the prediction accuracy approximately 34\% from the \TLS\ based models.
For near and far geometries the \GMR\ based models increased the prediction accuracy by 40\% from the \TLS\ based models.
The \TLSS\ based models increased the prediction accuracy approximately 29\% from the \TLS\ based models.

In Figure \ref{fig:both} we showed the log of the RMS prediction error for each model for the combination categories of geometries and frequencies.
For the same-same category the \GMRS\ based model got better prediction accuracy then the \GMR\ based model.
For the other categories, we obtained better prediction accuracy for predicting the closer gait to the model gait, for same frequency with near and far geometry and for the different frequency with same geometry.

\subsection{Independent Effects and Interaction Effects}

\newcommand{\xC}[1]{{\cdot\text{\bf {#1}}}}
We obtained a Generalized Linear Model (GLM) of the log RMS error, with single term direct effects:
\begin{align}
 y = & \text{\bf ~const.}+ a\xC{\SUDS}  + b\xC{\GMR} \nonumber\\
 & + c\xC{(model-period}<\text{\bf data-period)} \nonumber\\
 & + d\xC{(mode-period=data-period)} \\
 &+    e\xC{(model-gait = data-gait)} \nonumber\\
 & + f \xC{(model-gait far from data-gait)}\nonumber
\end{align}
We also obtained a GLM with all first-order interaction terms of the same predictors.
The results are in Table~\ref{tab:glm_compare}.

In the direct-effects model, both \SUDS\ and \GMR\ considerably reduce error, with \GMR\ showing a larger average effect.
Additionally, in-sample modeling of period and gait shape further decreased error, while significant gait mismatches increased it.
Interestingly, there was no direct cost or benefit to using data collected at lower frequencies for modeling motions at higher frequencies.

The full interaction model indicated that these effects has strong interaction terms.
The benefit of switching from \TLS\ to \TLSS\ was about 52\% error reduction, where the switch the \GMR\ would have given a 62\% reduction.
Apply both enhancements to get to \GMRS\ gives a 69\% reduction.

When using \SUDS\ or \GMR\ or both, almost all of the direct effect of frequency dependence was removed. 
Almost all the effects of gait were small, with model gait being far from training data gait being a consistent negative across all algorithm choices.

\begin{table*}
\centering
\caption{GLM coefficient comparison: direct effects model and first-order interactions model. %
  Because the dataset was large, all effects were statistically significant at better than $p<10^{-5}$, and we omitted the $p$ values from our report. %
  Instead, we tabulate the effects in order of decreasing magnitudes. %
  }
\label{tab:glm_compare}
\small
\begin{tabular}{lS[table-format=3.2]c}

\multicolumn{3}{c}{\large\bf Direct effects model}\\
Term & & Conf. Interval \\
\hline
Intercept
& 68.72 & $[68.63,\;68.80]$ \\

\GMR
& -24.31 & $[-24.38,\;-24.25]$ \\

\SUDS
& -16.61 & $[-16.67,\;-16.54]$ \\

model-period=data-period
& -8.34 & $[-8.42,\;-8.26]$ \\

model-gait far from data-gait
& 7.23 & $[7.15,\;7.30]$ \\

model-gait=data-gait
& -5.75 & $[-5.84,\;-5.65]$ \\

model-period$<$data-period
& -0.66 & $[-0.74,\;-0.58]$ \\
\hline
\\
\multicolumn{3}{c}{\large\bf First order interactions model} \\
Term &  & Conf. Interval \\
\hline
Intercept
& 91.36 & $[91.21,\;91.50]$ \\

\GMR
& -56.32 & $[-56.48,\;-56.16]$ \\

\SUDS
& -47.53 & $[-47.68,\;-47.37]$ \\

\SUDS$ \times$\GMR
& 40.36 & $[40.23,\;40.49]$ \\

model-period$<$data-period
& -19.80 & $[-19.98,\;-19.63]$ \\

model-period=data-period
& -18.98 & $[-19.15,\;-18.80]$ \\

\SUDS$ \times$(model-period$<$data-period)
& 17.34 & $[17.19,\;17.50]$ \\

\GMR$\times$(model-period$<$data-period)
& 15.99 & $[15.84,\;16.15]$ \\

\GMR$\times$(model-period=data-period)
& 10.11 & $[9.96,\;10.27]$ \\

\SUDS$ \times$(model-period=data-period)
& 8.52 & $[8.37,\;8.68]$ \\

\GMR$\times$(model-gait far from data-gait)
& 6.59 & $[6.45,\;6.74]$ \\

(model-period$<$data-period)$\times$(model-gait far from data-gait)
& 5.73 & $[5.55,\;5.90]$ \\

\SUDS$ \times$(model-gait far from data-gait)
& 4.52 & $[4.38,\;4.67]$ \\

(model-period=data-period)$\times$(model-gait far from data-gait)
& 4.01 & $[3.83,\;4.18]$ \\

model-gait=data-gait
& -4.07 & $[-4.27,\;-3.87]$ \\

(model-period=data-period)$\times$(model-gait=data-gait)
& -3.02 & $[-3.24,\;-2.80]$ \\

model-gait far from data-gait
& -1.58 & $[-1.74,\;-1.41]$ \\

(model-period$<$data-period)$\times$(model-gait=data-gait)
& -1.34 & $[-1.56,\;-1.12]$ \\

\SUDS$ \times$(model-gait=data-gait)
& -0.26 & $[-0.44,\;-0.07]$ \\

\GMR$\times$(model-gait=data-gait)
& -0.19 & $[-0.37,\;-0.01]$ \\
\hline
\end{tabular}

\end{table*}

\subsection{Influence of Dataset Size}

\begin{figure*}[h]
    \centering
    \includegraphics[trim={3cm 0 3cm 0 },clip,width=1\linewidth]{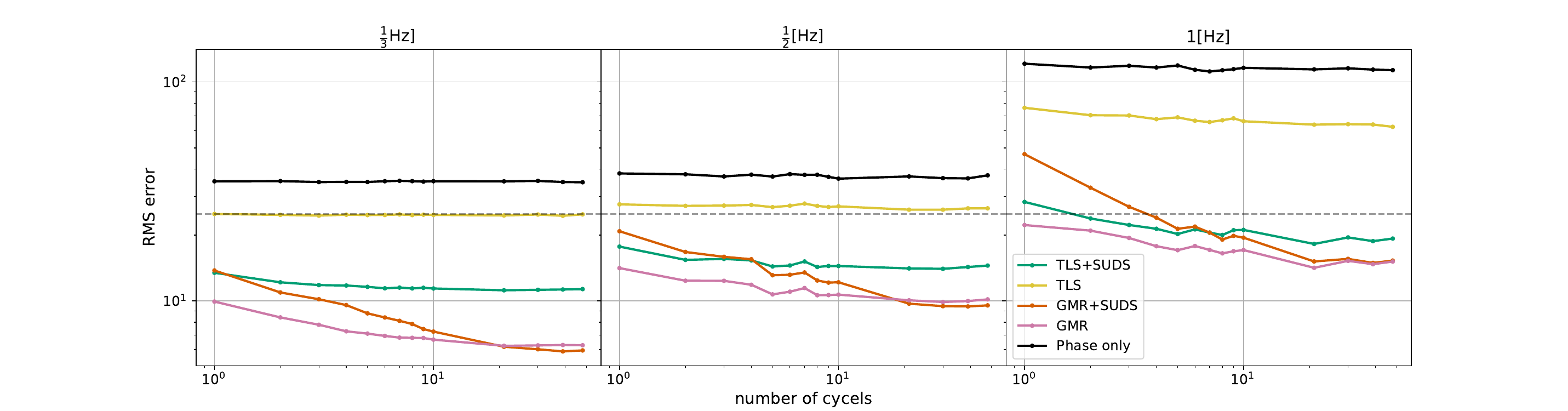}
    \caption{The  RMS error as a function of the number of cycles for three frequencies 1/3,1/2, and 1 [Hz], comparing five modeling approaches: \TLS, \TLSS, \GMR, \GMRS, and a phase-only baseline. %
    (note: axes are log-log) %
(for spread see Table \ref{tab:iqr}) %!!! check freq reasoning in discussion
}
    \label{fig:cycels}
\end{figure*}

We examined the dependency of log modeling error of on the log quantity of training data (see Fig.~\ref{fig:cycels}), to discern how dependent each model is on the availability of data.
As a reference, we also added a pure phase-based model -- the phase dependent constant term of the \TLS\ model.

Across all frequencies, increasing the number of cycles consistently reduced RMS error for all methods indicating improved estimation accuracy with additional data, although the effect was small for the \TLS\ and \TLSS\ models.
This reduction was most pronounced during the first 5–10 cycles, after which performance gradually saturated.
In all conditions, incorporating \SUDS\ resulted in improvements over the corresponding baseline models.
Specifically, \TLSS\ outperformed \TLS\ across the entire range of cycle counts, and \GMRS\ outperformed \GMR\ at high cycle counts.
This effect demonstrated that \SUDS\ improved sample efficiency and accelerated convergence.
Comparing models, \GMR-based approaches achieved substantially lower RMS error than \TLS-based approaches at all frequencies.
While \SUDS\ reduced the error floor for both models, \GMR\ remained superior, indicating that increased model expressiveness played a dominant role in achieving low prediction error.

The excitation frequency significantly affected performance.
Errors were highest at 1 [Hz], decreased at 1/2 [Hz], and were lowest at 1/3 [Hz] across all models.
This trend is common in many mechanical systems. 
At higher excitation frequencies and energies, more vibrational modes achieve an observable amplitude and the motion becomes more complex. 

The phase-only baseline demonstrated a high RMS error and showed little improvement as cycles or frequency increased.
This highlighted the limitations of relying solely on phase information and emphasized the need to understand fundamental geometric connection.

\section{Discussion}

In this paper we set up to explore the benefit of \GMR\ (A-GBR from \citet{hu2025learning}) and \GMRS\ over the previously published \TLS\ \citep{bittner2018geometrically} and \TLSS\ \citep{bittner2022data}.
Neither \TLSS\ nor \GMRS\ have previously been tested on data from physical robots.
In addition, we explored whether these modeling approaches could successfully model the motion of a robot with significant un-actuated dynamics, and underlying body-environment interaction physics which are likely not amenable to modeling with current tools from physics.

Our 4-flipper, 3-segment granular swimmer provided a dataset for exploring these questions in some detail.
It also allowed us to peer into the question of sample-efficiency of these methods -- how much training data is needed for each method, and how far would each method improve with more data.

Somewhat surprisingly, when trying to generalize to gaits with a different shape or frequency, the \GMR\ model which ignores the flippers out-performs the more elaborate \GMRS\ model (see Figs.~\ref{fig:freq},\ref{fig:gaits},\ref{fig:both}).

The picture becomes a bit clearer when looking at the GLM models that describe the impact of various factors on model quality.
Without interaction, adding \SUDS\ and switching from \TLS\ to \GMR\ lead to comparable single variable effects improving the outcome (see Table~\ref{tab:glm_compare}).
With interaction we see that the interaction term is fairly large and in the opposite sign.
This suggests that the change to \GMR\ captures some of the same information that the addition of \SUDS\ provided.
It seems that about 85\% of the contribution of the passive shape variables can be expressed through a non-linear dependence on $\dot r_a$. 
The similarity of the \SUDS\ and \GMR\ interaction terms with the other variables, i.e. the fact that the \SUDS$\times$(variable) coefficient is close to the \GMR$\times$(variable), is also consistent with the notion that they do capture similar features of the dynamics.

When we examined the dependency of the various models on the amount of available data (see Fig.~\ref{fig:cycels}), the results were fairly consistent.
The phase-based model gave the poorest model, and reached this level with a minuscule amount of data. 
The \TLS\ model improved on this by a factor of about $\times 3$ showing only a tiny improvement with more training data.
Adding \SUDS\ to \TLS\, along the lines of \citet{bittner2022data}, gave a $\times 3$ to $\times 10$ improvement.

For small training data-sets that consisted of only a few cycles of data, \TLSS\ was as good or better than \GMRS\, but \GMR\ was better than both of them.
However, \GMRS\ improved rapidly with training data.
In fact, in most of Fig.\ref{fig:cycels}, \GMRS\ improved at a log-log slope close to $-1/2$, which is the rate at which the standard error of IID Gaussian noise is reduced by averaging.
This is the expected signature of a model that captures ``all the physics'' so that the residual becomes uncorrelated white noise.

\subsection{Conclusions}

We have successfully demonstrated that adding  \SUDS\ terms from \citet{bittner2022data} to a \TLS\ model improves the ability to model a physical robot by a $\times 3$ to $\times 10$ improvement.
We have also shown that \GMR, or more specifically the A-GBR algorithm from \citet{hu2025learning}, is even more effective, but is out-competed by \GMRS\ when enough data is available.
The rate of convergence of \GMRS\ suggests that it may have captured all the available structure in the data.
Additionally, we have seen that learning the model using slower recordings is highly beneficial, and that extrapolating too far from a training gait comes with a moderate cost.

\subsection{Future Work}

Perhaps the most obvious question is how these algorithms compare with more traditional machine learning approaches.
The work in \cite{hu2025learning} suggests that \GMR\ requires $\times 10$ less data than a Multi-Layer Perceptron (MLP) modeling the motility map, placing MLP and \GMRS\ at similar data requirements.
How do they compare in practice? 
More generally, a survey comparison of traditional machine learning tools applied to this task would help the community choose the appropriate tools for the task.

For a practical robot experiencing a changing environment and potential damage and performance degradation of parts, an online version of \GMR\ or \GMRS\ would be highly beneficial. 
We might improve these online tools even more by noting that the transition from \GMR\ to \GMRS\ could be made more gradual, adding shape variables as more data becomes available. 
Given that our choice of state variables was derived from PCA it is possible that other data reduction techniques such as the use of auto-encoder latent states would give even better ``shape variable'' predictors. 

The classical motility map is linear in $\dot r$, but asymmetric friction can lead to a more complicated relationship which still gives rise to a motility map, albeit one which is merely continuous, homogeneous, and convex in $\dot r$ (a.k.a. as a ``Finsler metric''; see \citet{hatton2025asymmetric}).
Testing the ability of the \GMR\ and \GMRS\ model on systems with Finsler contact models may be another fruitful future direction.

Finally, note that the full reconstruction equation Eqn.~\ref{eqn:recon} contains a ``group momentum'' term $p$. 
For some systems where $p$ is significant, the dimension of $p$ might still be low. 
For example, almost all of a car's momentum is 1-dimensional along its the direction of motion; all other inertial motions are suppressed by tire traction and the suspension. 
Can these approaches for learning data-driven motility maps be extended to systems with low-dimensional momentum?

\subsection{Final notes}
We believe that the insights geometric mechanics provides into the physics of propulsion create a huge opportunity for improving our models of robot and animal locomotion. 
They also provide structure which can be exploited to create learning algorithms that may enable the robots of the future to move reliably and adapt quickly to changes in the body-environment interaction, and may provide insights into the evolution and the pathologies of animal locomotion control.

% ----------------------------
% Table 1: 1/3 Hz
% ----------------------------
\begin{table*}[ht]
\centering
\caption{Spread of results in Fig.\ref{fig:cycels}}\label{tab:iqr}
\label{tab:iqr_1_3Hz}
\scriptsize
\begin{tabular}{lrrrrrrr}

\\
\multicolumn{7}{c}{\large\bf Inter-quartile range (IQR) values at 1/3 Hz} \\
\\
\hline
Model & 1 & 2 & 3 & 4 & 5 & 6 & 7 \\
\hline
\TLSS\ & 6.12 & 5.57 & 5.40 & 5.35 & 5.31 & 5.20 & 5.27 \\
\GMRS\ & 5.82 & 4.69 & 4.32 & 4.16 & 3.77 & 3.62 & 3.50 \\
\TLS\  & 11.80 & 11.70 & 11.67 & 11.83 & 11.69 & 11.71 & 11.79 \\
\GMR\  & 4.30 & 3.53 & 3.28 & 3.10 & 3.02 & 2.94 & 2.91 \\
Phase only & 15.88 & 16.05 & 15.77 & 15.84 & 15.80 & 16.09 & 16.10 \\
\hline
\hline
Model & 8 & 9 & 10 & 21 & 30 & 39 & 48 \\
\hline
\TLSS\ & 5.26 & 5.28 & 5.22 & 5.12 & 5.13 & 5.22 & 5.22 \\
\GMRS\ & 3.38 & 3.23 & 3.10 & 2.65 & 2.59 & 2.51 & 2.53 \\
\TLS\  & 11.71 & 11.71 & 11.67 & 11.65 & 11.70 & 11.54 & 11.62 \\
\GMR\  & 2.85 & 2.83 & 2.81 & 2.63 & 2.62 & 2.62 & 2.61 \\
Phase only & 15.99 & 15.88 & 16.06 & 15.98 & 15.97 & 15.88 & 15.92 \\
\hline

\\
\multicolumn{7}{c}{\large\bf Inter-quartile range (IQR) values at 1/2 Hz} \\
\\
\hline
Model & 1 & 2 & 3 & 4 & 5 & 6 & 7 \\
\hline
\TLSS\ & 8.08 & 7.07 & 7.05 & 6.94 & 6.54 & 6.67 & 6.87 \\
\GMRS\ & 9.26 & 7.44 & 7.10 & 6.87 & 5.87 & 5.90 & 5.90 \\
\TLS\  & 12.71 & 12.61 & 12.72 & 12.77 & 12.43 & 12.61 & 12.95 \\
\GMR\  & 6.42 & 5.51 & 5.48 & 5.26 & 4.78 & 4.82 & 4.98 \\
Phase only & 16.56 & 16.37 & 15.96 & 16.47 & 15.99 & 16.25 & 16.11 \\
\hline
Model & 8 & 9 & 10 & 21 & 30 & 39 & 48 \\
\hline
\TLSS\ & 6.51 & 6.55 & 6.63 & 6.53 & 6.42 & 6.51 & 6.65 \\
\GMRS\ & 5.52 & 5.37 & 5.33 & 4.39 & 4.23 & 4.20 & 4.24 \\
\TLS\  & 12.72 & 12.50 & 12.55 & 12.16 & 12.13 & 12.29 & 12.32 \\
\GMR\  & 4.66 & 4.70 & 4.68 & 4.49 & 4.36 & 4.42 & 4.49 \\
Phase only & 16.26 & 15.96 & 15.54 & 16.18 & 15.79 & 15.72 & 16.15 \\
\hline

\\
\multicolumn{7}{c}{\large\bf Inter-quartile range (IQR) values at 1 Hz} \\
\\
\hline
Model & 1 & 2 & 3 & 4 & 5 & 6 & 7 \\
\hline
\TLSS\ & 13.18 & 11.07 & 10.38 & 9.97 & 9.41 & 9.83 & 9.56 \\
\GMRS\ & 20.77 & 15.12 & 12.54 & 11.14 & 9.90 & 10.07 & 9.42 \\
\TLS\  & 34.11 & 32.15 & 31.92 & 30.58 & 31.79 & 30.24 & 30.45 \\
\GMR\  & 10.46 & 9.71 & 8.97 & 8.26 & 7.93 & 8.28 & 7.94 \\
Phase only & 56.75 & 55.73 & 57.58 & 56.53 & 58.49 & 55.19 & 55.62 \\
\hline
Model & 8 & 9 & 10 & 21 & 30 & 39 & 48 \\
\hline
\TLSS\ & 9.27 & 9.76 & 9.77 & 8.49 & 9.04 & 8.81 & 9.13 \\
\GMRS\ & 8.88 & 9.29 & 8.92 & 7.06 & 7.20 & 6.93 & 7.13 \\
\TLS\  & 30.71 & 31.13 & 30.11 & 29.63 & 29.27 & 29.15 & 28.70 \\
\GMR\  & 7.68 & 7.85 & 7.89 & 6.51 & 7.02 & 6.90 & 7.15 \\
Phase only & 55.92 & 56.87 & 57.74 & 56.77 & 58.07 & 56.90 & 57.55 \\
\hline
\end{tabular}
\end{table*}

\FloatBarrier
% \bibliography{ref.bib}

\end{document}